\documentclass[runningheads]{llncs}

\usepackage{eccv}

\usepackage{eccvabbrv}

\usepackage{graphicx}
\usepackage{booktabs}
\newcommand{\minisection}[1]{\vspace{0.04in} \noindent {\bf #1}\ \ } 

\usepackage[accsupp]{axessibility}  

\usepackage{hyperref}

\usepackage{orcidlink}

\begin{document}

\title{Unleashing the Potential of Synthetic Images: A Study on Histopathology Image Classification} 

\titlerunning{Synthetic Image Potential in Histopathology Classification}

\author{Leire Benito-Del-Valle\inst{1}\orcidlink{0000-0002-2152-1281} \and
Aitor Alvarez-Gila\inst{1}\orcidlink{0000-0002-5955-3211} \and
Itziar Eguskiza\inst{1,2}\orcidlink{0009-0005-7559-2156} \and
Cristina L. Saratxaga\inst{1}\orcidlink{0000-0003-1429-7936}}

\authorrunning{L.~Benito-Del-Valle et al.}

\institute{TECNALIA, Basque Research and Technology Alliance (BRTA), Parque Tecnol\'{o}gico de Bizkaia, C/ Geldo. Edificio 700, E-48160 Derio - Bizkaia (Spain)
\email{\{leire.benitodelvalle,aitor.alvarez,itziar.egusquiza,cristina.lopez\}@tecnalia.com}\\
\and
University of the Basque Country, Plaza Torres Quevedo, 48013 Bilbao, Spain}

\maketitle

\begin{abstract}
  Histopathology image classification is crucial for the accurate identification and diagnosis of various diseases but requires large and diverse datasets. Obtaining such datasets, however, is often costly and time-consuming due to the need for expert annotations and ethical constraints. To address this, we examine the suitability of different generative models and image selection approaches to create realistic synthetic histopathology image patches conditioned on class labels. Our findings highlight the importance of selecting an appropriate generative model type and architecture to enhance performance. Our experiments over the PCam dataset show that diffusion models are effective for transfer learning, while GAN-generated samples are better suited for augmentation. Additionally, transformer-based generative models do not require image filtering, in contrast to those derived from Convolutional Neural Networks (CNNs), which benefit from realism score-based selection. Therefore, we show that synthetic images can effectively augment existing datasets, ultimately improving the performance of the downstream histopathology image classification task. 
   
  \keywords{Histopathology image classification \and Bioimage synthesis \and Bioimage data augmentation \and Diffusion probabilistic models \and Generative models}
\end{abstract}

\section{Introduction}
\label{sec:intro}
Histopathology requires the examination of tissues under a microscope to diagnose diseases, damages, or other abnormalities~\cite{basic_pathology}. The task of histopathology image classification consists of translating these visual examinations into meaningful, clinically accurate results, which provide precise and timely diagnosis, prognosis, and treatment for the patient. The diagnosis of histopathology images, which requires detecting subtle changes, is prone to inconsistencies among pathologists. This is why, over the past decade, computer-assisted diagnosis (CAD) algorithms have been developed to improve the efficiency and reliability of cancer care, complimenting the pathologist's perspective and providing quantitative and objective measurements of histopathologic features~\cite{bioengineering10111289,Gurcan2009-ig}.

These algorithms often require a large quantity of expert-annotated data to obtain high-level accuracy, which is difficult to find. The labeling process entails expertise and manual annotation, which are time-consuming, expensive, and depend on the pathologist's interpretation and the staining protocol. Moreover, accessing and collecting high quality digital histopathology images remains a significant obstacle, particularly in countries where digitalization of samples is limited. Therefore, obtaining a large and diverse dataset with reliable labels is difficult and often time-consuming.

To address these challenges, researchers have turned to synthetic data generation as a supplementary data source~\cite{Gonzales2023-mr}. Data synthesis techniques usually rely on generative models, such as Variational Autoencoders (VAEs)~\cite{kingma2022autoencoding} and Generative Adversarial Networks (GANs)~\cite{goodfellow2014generative}, which learn to mimic the distribution of the real data~\cite{kazeminia2019gans}. Recently, Denoising Diffusion Probabilistic Models (DDPM)~\cite{DBLP:journals/corr/abs-2006-11239} have emerged as a promising addition to the collection of generative models. They can capture intricate details and nuances present in the original data, thus generating high-fidelity synthetic data that outperforms previous models in terms of quality and diversity~\cite{ddpm_beat_gans,ddpm_medicine}.

Despite these advances, there are still several setbacks in using synthetic data to classify histopathology images. One main issue is ensuring the quality and realism of the generated images, as poor-quality synthetic data can adversely affect the performance of the model~\cite{budach2022effects}. In addition, it is crucial to have filtering strategies in place to eliminate low-quality or irrelevant synthetic images from the dataset~\cite{XUE2021101816}. Finally, there is little understanding on how to integrate synthetic data into the deep learning model training workflow and how to strike a balance between synthetic and real data.

The goal of this work is to explore the efficacy of generative models to expand training sets with high-quality synthetic samples for histopathology image classification. We evaluate these models on the quality of the generated images and the improvement of the predictive performance of the classification task.

The main contributions of this paper are as follows:
\begin{itemize}
\item We propose employing diffusion probabilistic models to generate synthetic histopathology images, assessing the effect of different backbones.
\item  To enhance the performance of a downstream task, two post-processing approaches based on the realism score metric are applied and evaluated. Both approaches automatically discard low-quality or underrepresented samples.
\item We evaluate the most adequate method to use the synthetic images to improve the classification task. We contemplate whether it is best to use the real and synthetic images together or separately while training the classification model.
\item We perform thorough experiments on a reduced version PCAM lymph node histopathology dataset~\cite{Veeling2018-qh}. Our proposed method enhances the classification performance, compared with baseline models and recent GAN-based synthetic data expansion. Our dataset and results are described in ~\cref{sec:results}.
\end{itemize}

\section{Related Work}

\subsection{Histopathology Image Classification}
Machine learning algorithms, specifically deep learning models, have led to a more accurate classification of histopathology images~\cite{Ahmed2022-ja}, particularly in Whole Slide Images (WSI), which are high resolution, and require large datasets. To tackle this, the sliding window approach or patch-level classification is usually applied~\cite{Hou2015PatchBasedCN,10.1001/jamanetworkopen.2019.14645}. With this method in use, the classification of each individual patch is performed, and these results are then combined to obtain the final label of the original WSI. Consequently, to provide reliable support for diagnostic decisions, accurate patch-level image classification is essential, building on the expertise of human pathologists.

In particular, for Metastatic Breast Cancer (MBC) detection in lymph node biopsies, numerous supervised solutions have been proposed~\cite{10.1001/jama.2017.14585}. Liu \etal~\cite{DBLP:journals/corr/LiuGNDKBVTNCHPS17} presented a deep Convolutional Neural Network (CNN) capable of automatically detecting and localizing metastasis, significantly reducing false negative rates. Wang \etal~\cite{wang2016deep} introduced a GoogLeNet-based deep learning method that considerably diminished the human error rate. 

Although success has been obtained by employing fully supervised learning methods, these require an extensive amount of expert annotation of lymph node sections. Given that the annotation process can be costly, time-consuming, and laborious, there can often be little to barely any labeled data accessible to supervised learning models.

\subsection{Conditional Image Synthesis}
Generative AI has seen remarkable advancements throughout the last few years, mainly in conditional image generation, with models like Conditional Variational Autoencoders  (CVAEs)~\cite{ramchandran2022learning} and Conditional Generative Adversarial Networks (cGANs)~\cite{mirza2014conditional}.

On the one hand, CVAEs add an additional layer of conditionality to the Variational Autoencoder's (VAE)~\cite{kingma2022autoencoding} fundamental structure. These improvements enable CVAEs to produce data that captures both the underlying probabilistic distribution and the traits or input variables it is conditioned on.

On the other hand, cGANs, an enhancement upon the original GANs~\cite{goodfellow2014generative}, integrate conditional information for the generation of structured data. The production of data that complies with these conditions is made possible by inserting criteria or labels that guide both the generator and the discriminator.

Recently, an innovative generative modeling approach known as \emph{denoising diffusion models}~\cite{DBLP:journals/corr/abs-2006-11239} have emerged, which prioritize controlled data transformation through iterative noise refinement. The capacity of diffusion models to produce high-quality samples, even outperforming previous generative models~\cite{dhariwal2021diffusion,rombach2022highresolution}, has attracted heaps of interest. \Cref{sec:denoising_diffusion_model} describes in further detail the principles of denoising diffusion models and how we have applied them.

The popularity of these generative models has already reached the medical domain, in particular histopathology, where some implementations have been presented using both GANs~\cite{gan_multi_attribute,HistoGAN} and DDPMs~\cite{ddpm_medicine,ddpm_beat_gans}. However, none of these works show the usefulness of the generated images to improve the performance of discriminative downstream tasks.

\subsection{Synthetic Data Augmentation}
Data augmentation is a vital technique in machine learning and computer vision, strategically used to overcome challenges posed by limited training data and aid in the generalization capacities of models. It involves artificially increasing the training dataset by applying various transformations to existing samples. This practice exposes the model to a wider scope of data instances, thus enhancing its robustness and reducing the risk of overfitting. Such augmentation efforts strengthen the model's proficiency in recognizing patterns in novel, unseen data.

Traditional data augmentation~\cite{perez2017effectiveness} has been essential to the consistent improvement of the predictive performance of discriminative deep learning models. These methods apply diverse geometric and photometric transformations such as rotation, flipping, scaling, color jitter, or translation to the original data, creating a more diverse and varied training set. Recent strides in augmentation techniques include refined strategies like CutMix~\cite{cutmix} and Mixup~\cite{mixupzhang2018}, which combine samples or features to generate new data instances. These approaches have proven vital in boosting model performance by enabling the assimilation of knowledge from interpolated data points, leading to a more general understanding of the underlying patterns.

In the hopes of finding a more effective data augmentation approach, researchers have turned to synthetic data. Synthetic data augmentation entails the creation of artificial samples that resemble the original data. It is especially beneficial when obtaining a large labeled dataset is challenging or economically impractical, which is more frequent in the medical image domain. Notable studies have explored the effectiveness of synthetic data in enhancing the performance of machine learning models from medical applications~\cite{jimaging9040081}. Frid-Adar \etal~\cite{Frid_Adar_2018} use cGAN-generated images to improve the CNN's ability to classify liver lesions. Xue \etal~\cite{XUE2021101816} present a synthetic augmentation framework for histopathology image classification that selectively adds new synthetic image patches generated by a proposed HistoGAN. Hossain \etal~\cite{Hossain2023} generate synthetic data with a GAN, feature-based filtering, and shape transformations to create realistic and diverse nuclei patches. Yu \etal~\cite{yu2023diffusionbased} introduce a diffusion-based augmentation method for nuclei segmentation. Ktena \etal~\cite{ktena_generative_2024} use conditional diffusion models across various medical image modalities to enhance the robustness and fairness of machine learning systems in healthcare applications. 

All in all, synthetic data augmentation has achieved promising results, but usually only a generative model architecture is analyzed and most works blindly add synthetic data to the original data. Few consider the architecture of the used generative model or study the best way to introduce these images into the downstream task's training workflow.

\section{Methodology}
In this section, we define the problem at hand, explain how the denoising diffusion models work and the used architecture, and define the employed image selection approaches.

\subsection{Problem Definition}
The objective of this paper is to explore the efficacy of diffusion probabilistic models to expand training sets with high-quality synthetic samples in order to improve the downstream predictive performance of histopathology image classification models. Synthesizing these pathology images, however, is a challenging task compared to typical images in other domains. Histopathology image synthesis requires generating realistic textures and colors, preserving accurate nuclei boundaries, and avoiding artifacts. Moreover, the generated images should not be just visually pleasing but should also improve the performance of downstream tasks, such as segmentation, classification, or detection.


\subsection{Denoising Diffusion Model}
\label{sec:denoising_diffusion_model}
Denoising diffusion models amount to two main processes: the forward diffusion process and the reverse denoising process. The former gradually adds noise to the input until it converges to a white noise distribution. The latter learns to generate data by denoising; thus, it transforms noise into data. 

\subsubsection{Forward Diffusion Process.} 
The forward diffusion process, defined by the conditional Gaussian diffusion kernel $q(x_t\vert x_{t-1})$ at timestep $t$, entails generating progressively noisier samples. Given an initial data point (i.e. real image) $x_0$, it aims to create a sequence of intermediate noisy samples $x_1,x_2,...,x_{T}$ that gradually approach the target distribution (Gaussian noise) $x_T \sim \mathcal{N}(0,1) $. The sample at any given timestep $x_t$ can be obtained through $x_0$ following~\cite{DBLP:journals/corr/abs-2006-11239}:

\begin{equation}
    x_t=\sqrt{\bar{\alpha}_t}x_0 + \sqrt{(1-\bar{\alpha}_t)} \varepsilon
\end{equation}
where $\varepsilon \sim \mathcal{N}(0,1)$ and $\bar{\alpha}_t=\prod_{s=1}^{t}(1-\beta_s)$. Moreover, $\beta_s$ represents the noise scheduler that controls the amount of noise added in each timestep.

\subsubsection{Reverse Denoising Process.}
The reverse denoising process, given by the parametric denoising distribution $p_{\theta}(x_{t-1}\vert x_t)$, aims to recover the original data point from a given noisy sample. Therefore, it is the reverse process of the aforementioned, $x_{T}\sim \mathcal{N}(0,1) \rightarrow x_0$. It consists of an iterative subtraction of the noise predicted by the neural network ($\varepsilon_\theta$), which can be represented as the latter~\cite{DBLP:journals/corr/abs-2006-11239}:

\begin{equation}
    x_{t-1}=\frac{1}{\sqrt{\alpha_t}} (x_t-\frac{\beta_t}{\sqrt{1-\bar{\alpha}_t}}\varepsilon_\theta(x_t,t)+\sigma_tz)
\end{equation}
where $z \sim \mathcal{N}(0,I)$. 

\subsubsection{Training Objective.}
Similar to the Variational Autoencoder~\cite{kingma2022autoencoding}, the goal is to optimize the variational upper bound on negative log likelihood as follows:
\begin{equation}
    \mathbb{E}\left[-\log p_{\theta}(\mathbf{x}_{0})\right]\leq\mathbb{E}_{q}\left[-\log{\frac{p_{\theta}(\mathbf{x}_{0:T})}{q(\mathbf{x}_{1:T}|\mathbf{x}_{0})}}\right]=:L
\end{equation}

This equation can be further simplified as shown by Sohl-Dickstein \etal~\cite{sohldickstein2015deep} and Ho \etal~\cite{DBLP:journals/corr/abs-2006-11239}:
\begin{align}
    L& = \mathbb{E}_{q}\left[\right.{D_{\mathrm{KL}}(q({\bf x}_{T}|{\bf x}_{0})\ \|p({\bf x}_{T}))}+\sum_{t>1}{D_{\mathrm{KL}}(q({\bf x}_{t-1}|{\bf x}_{t},{\bf x}_{0})\ \|p_{\theta}({\bf x}_{t-1}|{\bf x}_{t})}\\
    &-\log p_{\theta}(x_0|x_1)\left.\right]= \mathbb{E}_{q}\left[ L_T+\sum_{t>1}L_{t-1}+L_0\right]
\end{align}

Since both $q({\bf x}_{t-1}|{\bf x}_{t},{\bf x}_{0})$ and $p_{\theta}({\bf x}_{t-1}|{\bf x}_{t})$ are Normal distributions, the KL divergence can be parameterized as:
\begin{align}
        L_{t-1}& = D_{\mathrm{KL}}(q({\bf x}_{t-1}|{\bf x}_{t},{\bf x}_{0})\ \|p_{\theta}({\bf x}_{t-1}|{\bf x}_{t}))\\
        & = \mathbb{E}_{\mathbf{x}_{0} \sim q(\mathbf{x}_{0}),\epsilon\sim N(\mathbf{0},\mathbf{I})}\left[\right.{\frac{\beta_{t}^{2}}{2\sigma_{t}^{2}(1-\beta_{t})(1-{\bar{\alpha}}_{t})}}\\
        &||\epsilon-\epsilon_{\theta}({\sqrt{\bar{\alpha}_{t}}}\,\mathbf{x}_{0}+{\sqrt{1-{\bar{\alpha}}_{t}}}\,\epsilon,t)||^{2}\left.\right]+C
\end{align}

where $\lambda_t=\beta_{t}^{2}/(2\sigma_{t}^{2}(1-\beta_{t})(1-{\bar{\alpha}}_{t}))$. This time-dependent term assures that the training objective is properly weighted for the maximum data likelihood training, often being very large for small values of $t$. Ho \etal~\cite{DBLP:journals/corr/abs-2006-11239} discovered that setting $\lambda_t=1$ improves sample quality, and thus proposed a simplified loss:
\begin{align}
      L_{\mathrm{simple}}(\theta) & = \mathbb{E}_{t,\mathbf{x}_{0},\epsilon}\left[\|\epsilon-\epsilon_{\theta}(\sqrt{\overline{{{\alpha}}}_{t}}{\bf x}_{0}+\sqrt{1-\overline{{{\alpha}}}_{t}}\epsilon,t)\|^{2}\right] \\
          & = \mathbb{E}_{t,\mathbf{x}_{0},\epsilon}\left[\|\epsilon-\epsilon_{\theta}(x_t)\|^{2}\right]
\end{align}

Therefore, $L_{\mathrm{simple}}$ refers to a Mean-Squared Error (MSE) loss defined on the difference between the actual and estimated noise for the timestep $t$.

\subsubsection{Architecture.}
\label{sec:architecture}
\begin{figure}[tb]
  \centering
  \includegraphics[width=0.7\linewidth]{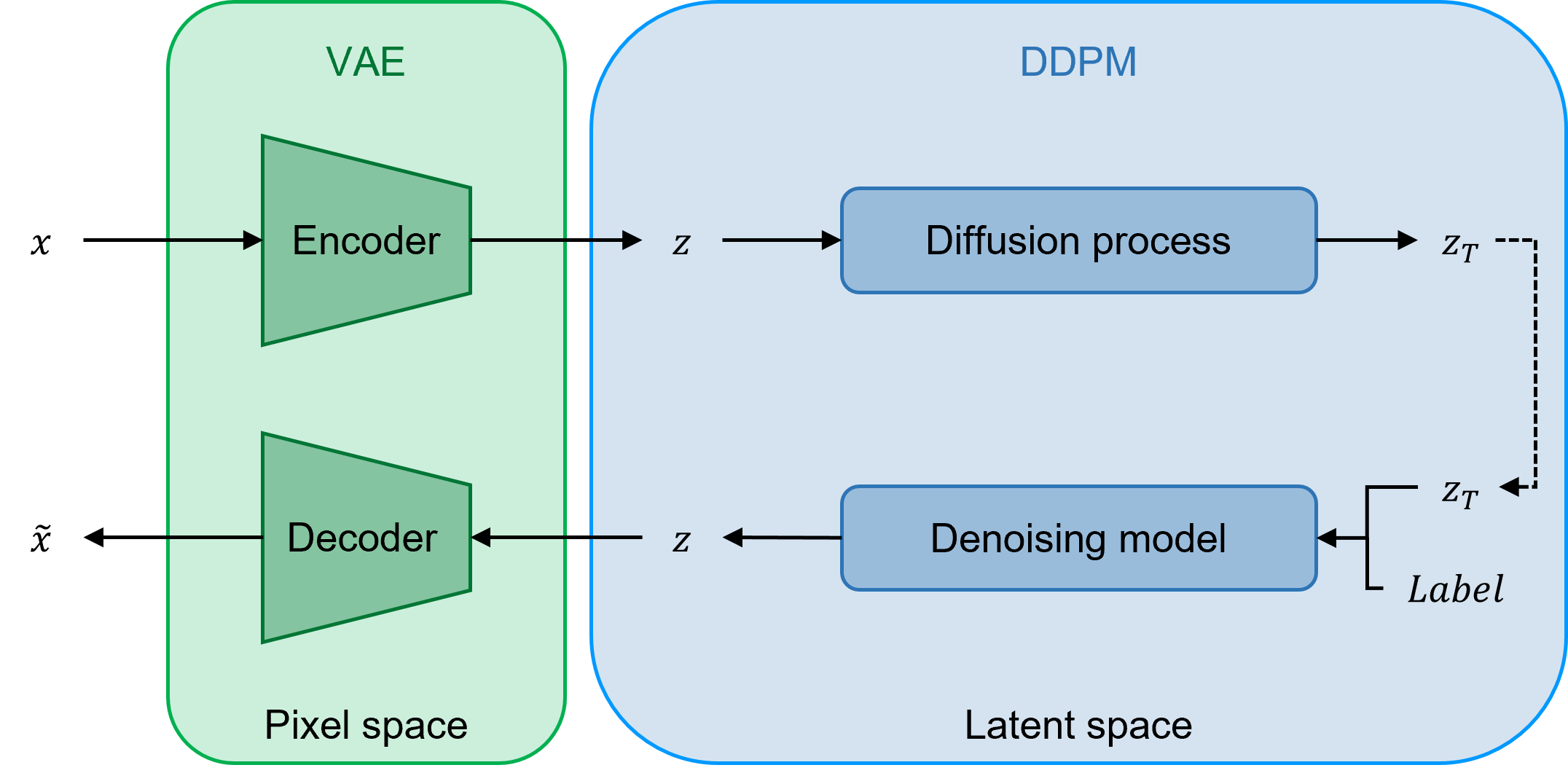}
  \caption{General overview of the latent diffusion model. First, the VAE converts the pixel space of the input image $x$ into a more compact spatial representation $z$. Next, we apply the diffusion model, which first employs the Gaussian diffusion process to transform the provided data distribution into a Gaussian distribution $z_T$. The denoising model is then trained to remove the added noise taking into account the class label. Finally, using the VAE decoder, the output of the diffusion model is converted back into pixel space.
  }
  \label{fig:dif_model}
\end{figure}

Inspired by Rombach \etal~\cite{rombach2022highresolution}, we opt for a latent diffusion model, as shown in ~\cref{fig:dif_model}. This type of models work on a more compact space, thus both the training and sampling time are drastically reduced. Therefore, the proposed model consists of two components: a VAE that will perform the bidirectional conversion between the pixel and the latent space and the diffusion model that is in charge of the generation, acting exclusively over the latent space. 

\begin{figure}[tb]
  \centering
  \begin{subfigure}{0.48\textwidth}
    \centering
    \includegraphics[width=0.4\linewidth]{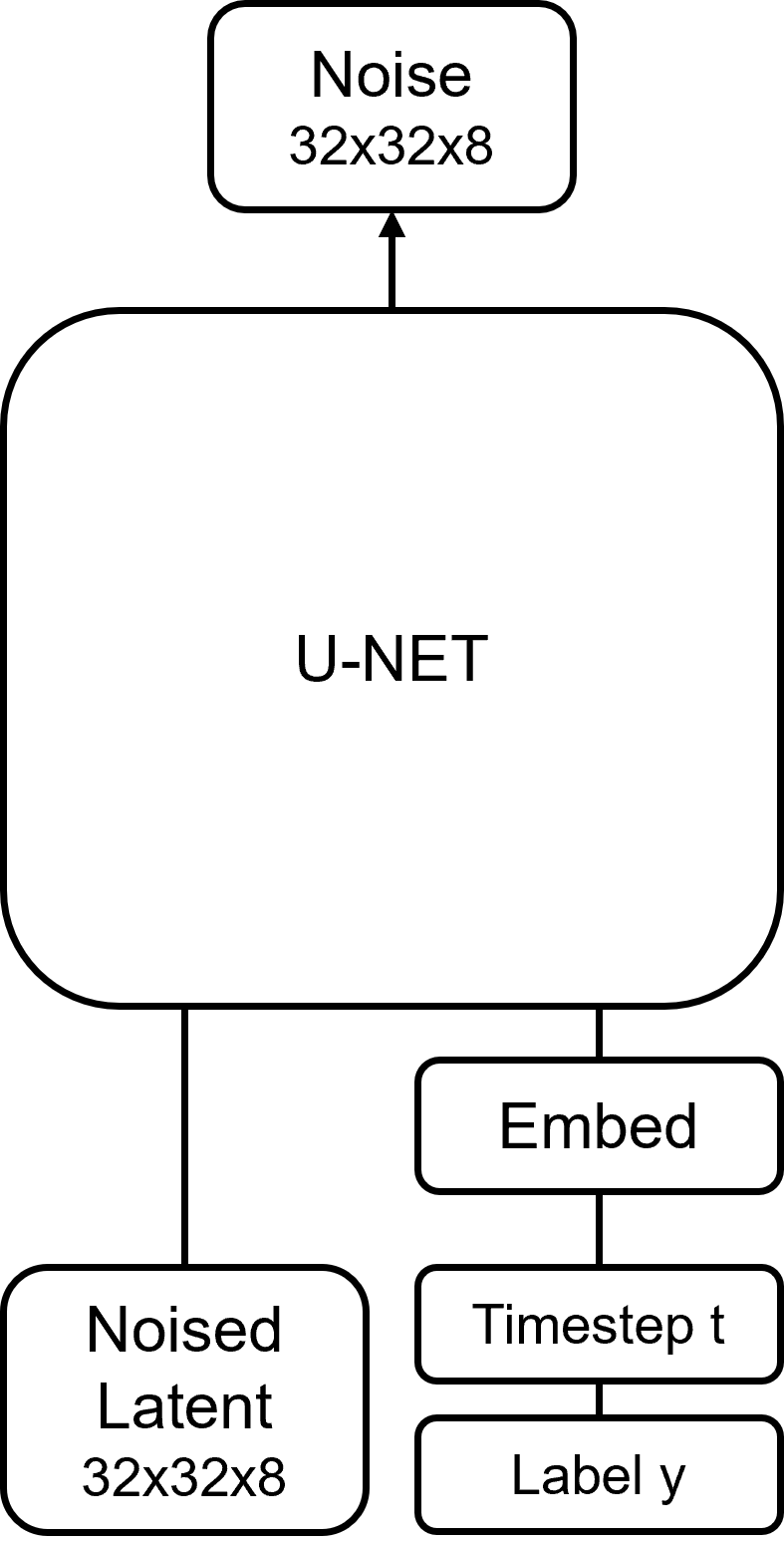}
    \caption{U-Net based diffusion model}
  \end{subfigure}
  \hfill
  \begin{subfigure}{0.48\textwidth}
    \centering
    \includegraphics[width=0.53\linewidth]{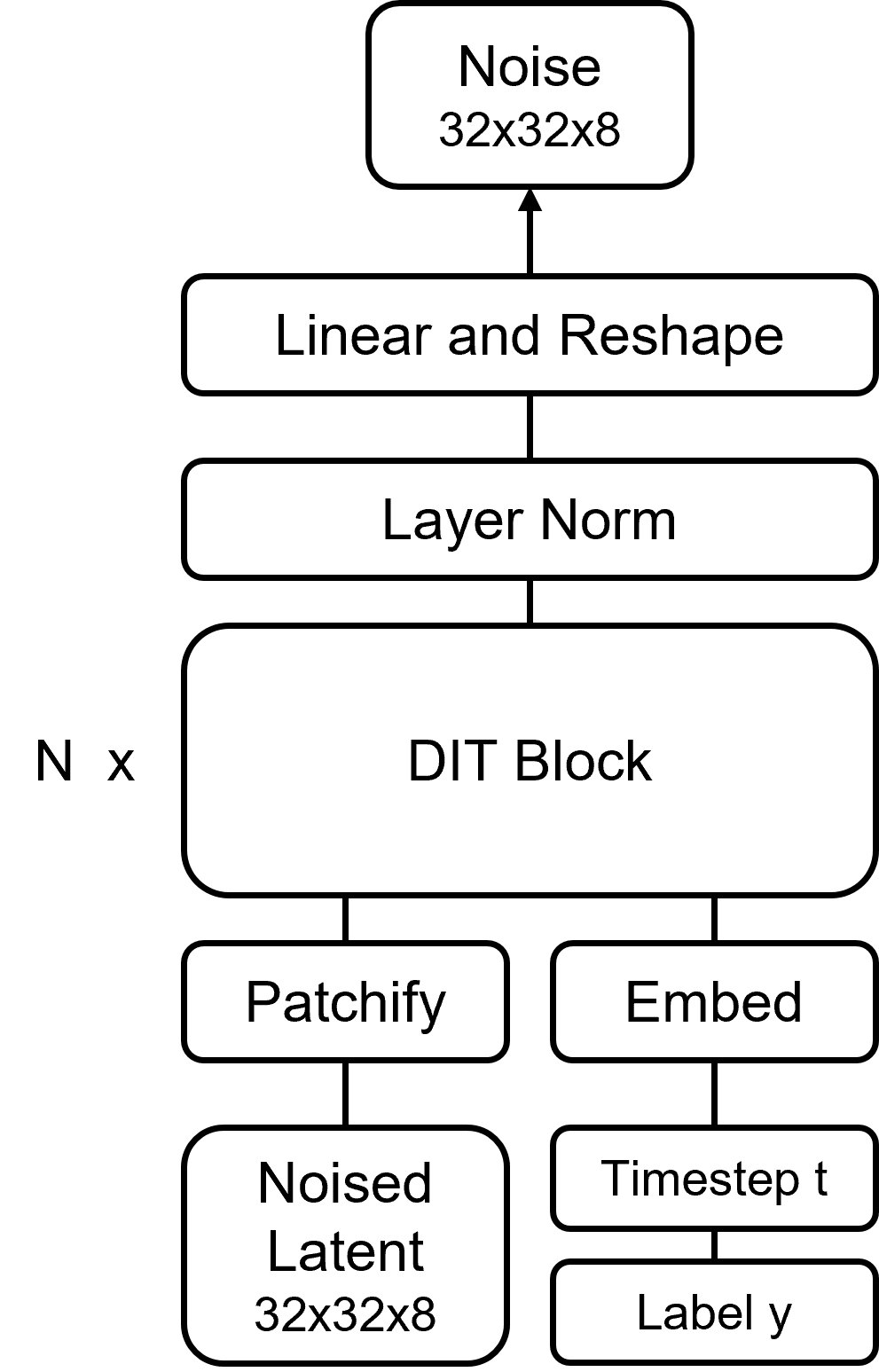}
    \caption{Transformer-based diffusion model}
  \end{subfigure}
  \caption{Architecture of the employed denoising models (\emph{U-Net based and transformer-based diffusion models}). Moreover, this image reflects how the conditioning is performed. The label is combined with the timestep, and after embedding them, they are inserted to the model.}
\label{fig:denoising_model}
\end{figure}

Moreover, the diffusion model can be further divided into two steps: the diffusion process and the denoising model. The former consists on iteratively adding noise to the input image following a pattern; in this case, a cosine scheduler was applied. The latter attempts to remove the added noise taking into account the class label. There are several backbones that can be used for this denoising model but they can be roughly divided into two main groups: U-Net~\cite{ronneberger2015unet}-based and Transformer-based~\cite{peebles2023scalable} diffusion models. A more detailed representation of the architecture of both these models is presented in ~\cref{fig:denoising_model}.

\subsection{Image Selection}
Although generative models have come to great lengths in creating realistic synthetic samples, they still fall behind in the produced quality and, in particular, diversity~\cite{mo2019mining,lee2021selfdiagnosing}, often generating data similar to the one they were trained on. Therefore, we opt for filtering the samples during the generation to obtain high-quality data. We use a realism-based method~\cite{kynkäänniemi2019improved}, which measures the similarity between a synthetic sample and the real data as follows:
\begin{equation}
        R(\phi_{g},\Phi_{r})=\operatorname*{max}_{\phi_{r}}\left\{{\frac{\|\phi_{r}-\mathrm{NN}_{k}\,(\phi_{r},\Phi_{r})\|_{2}}{\|\phi_{g}-\phi_{r}\|_{2}}}\right\}
\end{equation}
where, $\phi_{g}$ is the feature vector of a generated image and $\phi_{r}$ the feature vector of a real image from set $\Phi_{r}$. These features are obtained by a ResNet18~\cite{He_2016_CVPR} trained with the real data. In addition, the $\Phi_{r}$ consists of all the real training data or the data corresponding to one of the classes, depending on the approach (realism score or class-based realism score).

Under the standard realism score-based filtering scheme, a synthetic image is considered realistic if $R\geq 1$, which means that its feature vector $\phi_{g}$ is within the neighboring hypersphere of at least one real feature vector $\phi_{r}$. Therefore, whilst generating synthetic images those with $R < 1$ will be discarded.

Meanwhile, the class-based realism score presents a slight change to this method. In this approach, the realism score will be calculated per class, as the feature vectors will be obtained independently. This way, on top of filtering the low-quality samples, indirectly, those with incorrect labels will also be eliminated.

\section{Experiments}

\subsection{Dataset}
Our experiments employ the publicly available PatchCamelyon (PCam) benchmark dataset~\cite{Veeling2018-qh}, consisting of 327,680 color images from histopathologic scans of lymph sections, with a resolution of $96\times96$ pixels. Every example in the dataset is annotated with a binary label referencing the presence or absence of metastatic tissue. The dataset is divided into a training set containing 262,144 images and a validation and test set of 32,768 each, applying a hard-negative mining regime. To replicate a situation where limited training data is accessible, 5 non-overlapping subsets, each containing $10\%$ of the training set images (26,215) were randomly selected, ensuring each subset maintained the original class balance. These subsets were then used to train both the generative models and the classifier for the downstream task. Furthermore, for the classification task the whole validation and test sets were employed.

\subsection{Implementation Details}
As stated in \cref{sec:architecture}, for the image generation, a latent diffusion model is utilized. This model consists of two components, the VAE and the DDPM, which are each trained separately but with the same data (the VAE without labels and DDPM with them for the conditioning). Both models were trained for 1000 epochs with a $96\times96$ image resolution, 0.0001 learning rate and a batch size of 64. The only difference is the optimizer: we used Adam for the VAE and AdamW for the DDPM. Moreover, although two backbones were used for the diffusion model (\cref{fig:denoising_model}), both were trained with the previously stated setup. Furthermore, this training setup was executed five times, once for each of the five training subsets, and their results averaged.

For histopathology image classification, a ResNet18 model pre-trained on ImageNet~\cite{imagenet} was utilized for transfer learning. All experimental setups were trained for 100 epochs with the SGD (Stochastic Gradient Descent) optimizer and a momentum of 0.9. The learning rate and batch size were set as 0.0001 and 32, respectively. To train the model, except for the baseline, conventional data augmentation methods (i.e. random horizontal flip, random vertical flip, random rotation, and color jitter), were applied to enhance the model's generalization properties. To further improve the classification results, two approaches were employed that make use of the generated synthetic images:

\minisection{Augmentation:} the synthetic data is employed to expand the training dataset, thus, the real and synthetic data are randomly mixed and used together. 

\minisection{Transfer learning:} the model is first trained using the synthetic data as the training set, and then, starting from those weights, trained only with the real data.
  
It is worth mentioning that, in all the approaches, the validation and test sets are the same and are composed of real data with the original, human-labeled annotations.

All the experiments were implemented with Pytorch (version 1.13), PyTorch Lightning (version 1.6.0), and torchvision (version 0.14.0). The study was performed on an Ubuntu 20.04 server with an NVIDIA TITAN X GPU.

\subsection{Performance Evaluation Metrics}

\subsubsection{Image Quality Evaluation Metrics.}
The evaluation of synthetic images obtained through generative models is a complicated task and still an open research problem~\cite{betzalel2022study}. Usually, these models are assessed following two criteria: image quality (also known as image fidelity) and image diversity. While fidelity expresses the resemblance of the synthetic image to a real sample, diversity measures how varied the generated images are. In this study, some of the most widely used metrics are utilized: Improved Precision and Recall~\cite{kynkäänniemi2019improved}, a proposed Improved F1-score, and Fr\'{e}chet Inception Distance (FID)~\cite{heusel2018gans} which are described below.

\noindent\emph{Improved Precision and Recall} were proposed by Kynkäänniemi \etal~\cite{kynkäänniemi2019improved}, who assured that both the quality and distribution coverage of the produced samples are vital to evaluate a generative model. The \emph{Improved Precision} measures the fidelity as the percentage of produced samples inside the real data manifold. Whereas the \emph{Improved Recall} evaluates the diversity through the ratio of real samples located in the synthetic manifold. 


\noindent\emph{Improved F1-score} is a proposed metric inspired by the improved precision and recall. It combines the improved precision and recall into a single metric, since it is key to have both diverse and high-quality data.
\begin{equation}
    \mathrm{Improved F1\textnormal{-}score}= 2 \cdot \frac{\mathrm{Improved Precision} \cdot \mathrm{Improved Recall}}{\mathrm{Improved Precision} + \mathrm{Improved Recall}}
\end{equation}

\noindent\emph{Fr\'{e}chet Inception Distance (FID)} measures the agreement of the real and synthetic images by comparing the distribution of the generated images with the real images' distribution in the Inception-V3 latent space~\cite{kynkäänniemi2023role}. The lower the FID score, the better the quality of the generated images and the more similar they are to the real samples. The real and synthetic images are fed into the inception V3 model, and FID compares the mean and standard deviation of the features extracted from the last pooling layer, as given by:
\begin{equation}
    \mathrm{FID}=||\mu_{r}-\mu_{g}||^{2}+\mathrm{T_r}(\Sigma_{r}+\Sigma_{g}-2(\Sigma_{r}\Sigma_{g})^{1/2})
\end{equation}
where $\mu_{r}$ and $\mu_{g}$ are the mean of the real and synthetic samples' embeddings. Furthermore, $\Sigma_{r}$ and $\Sigma_{g}$ refer to their covariance.

\begin{figure}[tb]
\centering
    \begin{tabular}{ccccccc}
          \vspace{0.3em}
          &\multicolumn{2}{c}{\textbf{None}} & \multicolumn{2}{c}{\textbf{Realism score}} & \multicolumn{2}{c}{\textbf{Class-based RS}} \\
          \vspace{0.5em}
         \rotatebox[origin=c]{90}{\textbf{GAN}}&
        \raisebox{-0.5\height}{\includegraphics[width=0.14\textwidth]{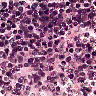}} &
        \raisebox{-0.5\height}{\includegraphics[width=0.14\textwidth]{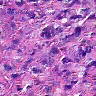}} &
        \raisebox{-0.5\height}{\includegraphics[width=0.14\textwidth]{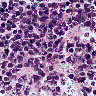}} &
        \raisebox{-0.5\height}{\includegraphics[width=0.14\textwidth]{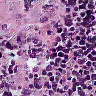}} &
        \raisebox{-0.5\height}{\includegraphics[width=0.14\textwidth]{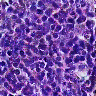}} &
        \raisebox{-0.5\height}{\includegraphics[width=0.14\textwidth]{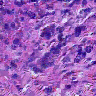}} \\
        \vspace{0.5em}
         \rotatebox[origin=c]{90}{\textbf{U-Net}}&
        \raisebox{-0.5\height}{\includegraphics[width=0.14\textwidth]{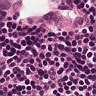}} &
        \raisebox{-0.5\height}{\includegraphics[width=0.14\textwidth]{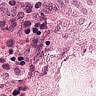}} &
        \raisebox{-0.5\height}{\includegraphics[width=0.14\textwidth]{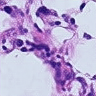}} &
        \raisebox{-0.5\height}{\includegraphics[width=0.14\textwidth]{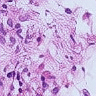}} &
        \raisebox{-0.5\height}{\includegraphics[width=0.14\textwidth]{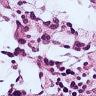}} &
        \raisebox{-0.5\height}{\includegraphics[width=0.14\textwidth]{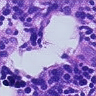}} \\
         \rotatebox[origin=c]{90}{\textbf{DiT}}&
        \raisebox{-0.5\height}{\includegraphics[width=0.14\textwidth]{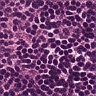}} &
        \raisebox{-0.5\height}{\includegraphics[width=0.14\textwidth]{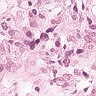}} &
        \raisebox{-0.5\height}{\includegraphics[width=0.14\textwidth]{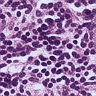}} &
        \raisebox{-0.5\height}{\includegraphics[width=0.14\textwidth]{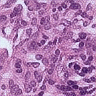}} &
        \raisebox{-0.5\height}{\includegraphics[width=0.14\textwidth]{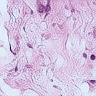}} &
        \raisebox{-0.5\height}{\includegraphics[width=0.14\textwidth]{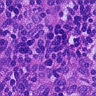}} \\
    \end{tabular}
\caption{Selection of generated patches with HistoGAN (\emph{GAN})  and both DDPM models: U-Net-based (\emph{U-Net}) and Transformer-based (\emph{DiT}). All these models were trained with the same dataset. Each dual column together represent the image selection approach used, namely none, realism score, and class-based realism score. In addition, these columns contain an image with the absence (odd) and presence (even) of metastatic tissue, respectively. All the generated synthetic images can be found here: \url{https://github.com/LeireBV/Synthetic_histopathology_dataset}}
\label{fig:gen_images}
\end{figure}

\subsubsection{Classifier Evaluation Metrics.}
To evaluate the performance of the histopathology classification algorithm, common metrics for image classification, such as accuracy, Area Under the ROC Curve (AUC), sensitivity, and specificity, are employed.

\section{Results and Discussion}
\label{sec:results}

This section assess the performance of the proposed approach. We compare the quality of the generated synthetic images against a recent GAN-based generative alternative and evaluate their utility in training downstream classification models. Notably, in all cases, we ensure dataset balance by using equal amounts of real and synthetic training data, even when filtering synthetic images, where we add additional filtered images to match the real data quantity.

\subsection{Image Quality Evaluation} 
\label{sec:image_quality_evaluation}

The objective of this experiment is to compare and contrast the quality of the synthesized images by our diffusion models against a generative adversarial network for histopathology, HistoGAN~\cite{HistoGAN}.  We chose this model because it is the most recent GAN implementation for the generation of histopathology images conditioned on the class. In ~\cref{fig:gen_images} we present samples of synthetic images generated by the three models using each of the three considered image selection approaches. The samples show that diffusion models, independent of the backbone, generate higher quality and more diverse samples.

\begin{table}[tb]
\caption{Image quality evaluation of the generative models.}
\label{tab:qualitative_results}
\centering
\begin{tabular}{@{}lllll@{}}
    \toprule
    Model & FID $\downarrow$ & \begin{tabular}{@{}l@{}}Improved\\ precision($\%$) $\uparrow$\end{tabular} & \begin{tabular}{@{}l@{}}Improved\\ recall($\%$) $\uparrow$\end{tabular} & \begin{tabular}{@{}l@{}}Improved\\ F1-score($\%$) $\uparrow$\end{tabular} \\
    \midrule
    \multicolumn{5}{l}{No image selection}\\
    \midrule
    HistoGAN & $126.71 \pm 17.13$ & $11.60 \pm 3.03$ & $0.00 \pm 0.00$ & $0.00 \pm 0.00$ \\
    U-Net based DDPM & $49.08 \pm 5.96 $ & $43.70 \pm 1.78 $ & $6.31 \pm 1.13$ & $10.98 \pm 1.70 $\\
    Transformer based DDPM & $\mathbf{35.70 \pm 2.55}$ & $\mathbf{45.17 \pm 2.67}$ & $\mathbf{11.17 \pm 0.55}$ & $\mathbf{17.9 \pm 0.72} $ \\
    \midrule
    \multicolumn{5}{l}{Realism score image selection}\\
    \midrule
    HistoGAN & $132.80 \pm 14.64$ & $8.53 \pm 3.04$ & $0.00 \pm 0.00$ & $0.00 \pm 0.00$ \\
    U-Net based DDPM & $51.94 \pm 6.07 $ & $42.27 \pm 2.41 $ & $3.42 \pm 0.63$ & $6.32 \pm 1.09 $\\
    Transformer based DDPM & $40.02 \pm 1.78$ & $41.08 \pm 2.10$ & $6.83 \pm 0.28$ & $11.72 \pm 0.49 $ \\
    \midrule
    \multicolumn{5}{l}{Class-based realism score image selection}\\
    \midrule
    HistoGAN & $129.87 \pm 13.65$ & $8.69 \pm 3.48$ & $0.02 \pm 0.04$ & $0.04 \pm 0.08$ \\
    U-Net based DDPM & $52.36 \pm 5.83 $ & $41.94 \pm 2.40 $ & $3.44 \pm 0.52$ & $6.35 \pm 0.90 $\\
    Transformer based DDPM & $40.34  \pm 1.78$ & $40.99 \pm 2.01$ & $6.85 \pm 0.38$ & $11.74 \pm 0.62 $ \\
    \bottomrule
\end{tabular}
\end{table}

\begin{figure}[tb]
  \centering
  \includegraphics[width=0.75\linewidth]{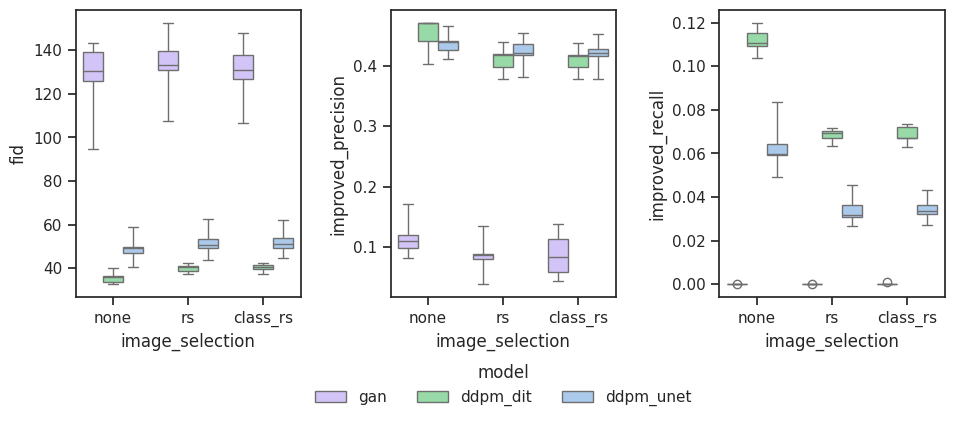}
   \caption{Image quality analysis. Each of the three graphs display the value of an image quality evaluation metric, comparing them depending on the generative model employed and the image selection approach.}
\label{fig:qualitative_analysis}
\end{figure}

This statement is supported by the quality metrics represented in ~\cref{fig:qualitative_analysis}, which have been obtained through the values in ~\cref{tab:qualitative_results}. Overall, the diffusion models noticeably outperform the HistoGAN. This difference is especially prominent in the recall, which always has a null value for the HistoGAN. This result is to be expected, as GANs are known to struggle with mode collapse~\cite{gan_mode}, namely the model tends to generate the same samples. Although there is little difference when changing the backbone of the diffusion model, the transformer-based approach generates slightly higher quality and more diverse samples.

It is worth mentioning that, even though the images were filtered to improve the resulting set, when doing so, the image quality metrics worsen. This decrease in performance is not critical, as the main reason for opting to apply image selection techniques is to improve the results of the downstream selection task.

\subsection{Evaluation on Downstream Classification Task} 
\begin{figure}[tb]
  \centering
  \includegraphics[width=0.85\linewidth]{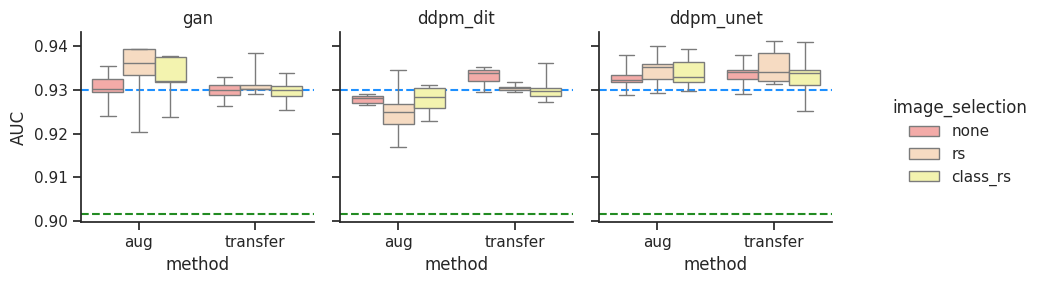}
   \caption{Analysis on the classification task performance. Each of the three graphs display the classification results (AUC) obtained with each generative model employed, comparing them depending on the training and the image selection approach. The baseline represents the result obtained when training the model with real data and no data augmentation (\emph{green line}). Traditional augmentation depicts the AUC when training the classification model with only real data and traditional data augmentation techniques (\emph{blue line}).}
\label{fig:quant_hue_method}
\end{figure}

The purpose of this experiment is to evaluate the use of synthetic images to improve the classification results. In ~\cref{tab:quantitative_results}, we report the scores obtained in a downstream binary classification task for the Resnet18 model trained over each of the selected 5 subsets, comparing the baseline, traditional data augmentation, and synthetic data augmentation using various generative models, training approaches, and image selections (\cref{fig:quant_hue_method}).

\begin{table}[tb]
\caption{Classification results of the baseline and augmentation model with diverse settings.}
\label{tab:quantitative_results}
\centering
\resizebox{\textwidth}{!}{%
\begin{tabular}{@{}lllll@{}}
    \toprule
    ResNet18 & Accuracy($\%$) $\uparrow$ & AUC($\%$) $\uparrow$ & Sensitivity($\%$) $\uparrow$ & Specificity($\%$) $\uparrow$ \\
\midrule
Baseline & $82.16 \pm 0.23$ & $90.17 \pm 0.20$& $81.32 \pm 1.53$  & $82.99 \pm 1.42$ \\
Traditional Augmentation  & $85.05 \pm 0.43$ & $92.99 \pm 0.26$ & $81.15 \pm 1.50$ & $88.95 \pm 1.13$ \\
\midrule
\multicolumn{5}{l}{No image selection}\\
\midrule
GAN Augmentation & $84.47 \pm 0.49 $ & $93.02 \pm 0.37$ & $76.35 \pm 1.2$ & $92.59\pm 0.39$ \\
DDPM   Unet   Augmentation & $85.22  \pm 0.39$ & $93.28  \pm 0.30$ & $81.24  \pm 1.13$ & $ 89.19  \pm 1.11$ \\
DDPM DiT   Augmentation & $84.95 \pm 0.74$ & $92.86  \pm 0.49$ & $81.32  \pm 1.82$ & $88.58 \pm 0.49$ \\
GAN transfer learning & $85.03 \pm 0.48$ & $92.98 \pm 0.22$ & $80.84 \pm 1.65$ & $89.21  \pm 0.81$ \\
DDPM Unet   transfer learning & $85.47\pm 0.36$ & $93.37 \pm 0.29$ & $81.40  \pm 1.11$ & $89.54  \pm 0.58$ \\
DDPM DiT transfer learning & $85.58\pm 0.57$ & $93.32\pm 0.32$ & $81.79 \pm 0.80$ & $89.38 \pm 1.02$ \\
\midrule
\multicolumn{5}{l}{Realism score image selection}\\
\midrule
GAN Augmentation & $85.19 \pm  0.94$ & $93.37 \pm 0.71$ & $77.44 \pm 2.26$ & $92.94 \pm 1.09$ \\
DDPM   Unet   Augmentation & $85.78  \pm 0.62$ & $93.45 \pm 0.36$ & $82.23 \pm 1.85$ & $89.33 \pm 0.72$ \\
DDPM DiT   Augmentation & $84.74  \pm 0.79$ & $92.51 \pm 0.58$ & $80.08 \pm 1.41$ & $89.40 \pm 0.55$ \\
GAN transfer learning & $85.38 \pm 0.56 $ & $93.18 \pm 0.34 $ & $82.10 \pm 1.48$ & $88.66 \pm 0.98$ \\
DDPM Unet   transfer learning & $\mathbf{85.87 \pm 0.64}$ & $\mathbf{93.54 \pm 0.38}$ & $\mathbf{83.30 \pm 1.80}$ & $88.45 \pm 0.62$ \\
DDPM DiT transfer learning & $85.06 \pm 0.38$ & $93.04 \pm 0.08$ & $80.68 \pm 1.41$ & $89.44 \pm 0.73$ \\
\midrule
\multicolumn{5}{l}{Class-based realism score image selection}\\
\midrule
GAN Augmentation & $84.82 \pm 0.85$ & $93.20  \pm 0.51 $ & $76.44 \pm 2.71$ & $\mathbf{93.2 \pm 1.17 }$ \\
DDPM   Unet   Augmentation & $85.64 \pm 0.65$ & $93.40 \pm 0.33$ & $82.20 \pm 2.48$ & $89.07 \pm 1.42$ \\
DDPM DiT   Augmentation & $84.98 \pm 0.43$ & $92.77 \pm 0.30$ & $82.69 \pm 2.15$ & $ 87.27\pm 1.46$ \\
GAN transfer learning & $85.00 \pm 0.53$ & $92.97 \pm 0.28$ & $80.60  \pm 2.00$ & $89.40 \pm 1.22$ \\
DDPM Unet   transfer learning & $85.57 \pm 0.65$ & $93.31 \pm 0.51$ & $82.12 \pm 0.77$ & $89.02 \pm 0.95$ \\
DDPM DiT transfer learning & $85.07 \pm 0.50$ & $93.04 \pm 0.30$ & $81.19 \pm 0.72$ & $88.94 \pm 0.63$ \\
\bottomrule
\end{tabular}
}
\end{table}

It is apparent that the U-Net based DDPM is the only generative model that consistently improves results when adding synthetic data, regardless of the training or image selection approach used, thus being the most stable and consistent.

\begin{figure}[tb]
	\centering
        \includegraphics[width=0.52\linewidth]{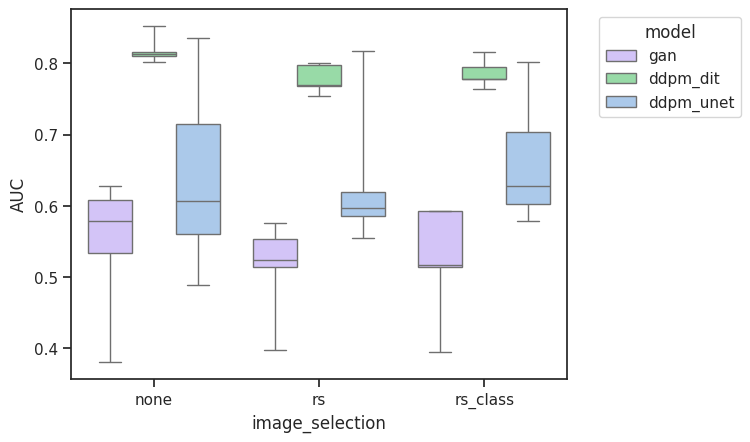}
   \caption{Analysis on the classification task performance only using synthetic images as the training data. The graph displays the classification results (AUC) obtained when using only synthetic data for the training, comparing them depending on the generative model and the image selection approach. }
\label{fig:synth_classification}
\end{figure}

As a rule of thumb, ~\cref{fig:quant_hue_method} evidences that diffusion model generated images yield better results when trained independently and then transfer the knowledge (transfer learning via fine-tuning), whereas HistoGAN samples need to be used together with real data (augmentation) to have a positive impact on the predictive results. This is evident when contrasting the transformer-based DDPM and the GAN: when augmentation is applied, the GAN produces better results, but in the transfer learning-based setup, the roles are reversed. We hypothesize that this phenomenon is closely related to the performance of the classifier trained only on synthetic data, as these are the weights used for transfer learning. Thus, in ~\cref{fig:synth_classification} we present the performance of the model when only synthetic images are used as the training dataset. It can be observed that the GAN model has comparatively lower results, which can justify the observed performance drop when transfer learning is applied. On the contrary, the complete opposite occurs with the transformer-based DDPM. As exhibited on ~\cref{fig:quant_hue_method} the transformer-based DDPM presents an improvement in the classification performance, most likely as a result of its exceptional results in the synthetic-only setup (\cref{fig:synth_classification}).

Furthermore, we can detect a correlation between the image quality metrics and the classification results. As stated in ~\cref{sec:image_quality_evaluation}, the best synthetic images are generated by the transformer-based DDPM, followed by the U-Net based one and the HistoGAN. Such behavior is also observed in ~\cref{fig:synth_classification}. This suggests that image quality metrics can predict which images are most useful for downstream tasks, but only when used alone, not in combination with real data.

In addition, we can observe that the performance of image filtering techniques depends on the architecture of the generative model used to produce the images. We can consider that, for this data domain, for CNN-based models (GAN and U-Net based DDPM) images should be filtered with the realism score, while for transformer-based models (transformer based DDPM) using unfiltered images is best.

All in all, the best implementation for each individual model, in no particular order, is as follows:  HistoGAN with augmentation and realism score selection, U-Net based DPPM with transfer learning and realism score selection, and transformer based DPPM with transfer learning and no selection. Overall, the U-Net based DDPM is clearly the best option to improve the results of a downstream classification task, as it always improves the baseline, no matter the setup.

\section{Conclusion}
In this paper, we evaluated the quality and utility of synthetic data as a data expansion technique for downstream task improvement, in the context of histopathology image classification. Results suggest that high-quality synthetic data does not ensure an improvement in a downstream task. When only synthetic samples are used during training, those with the best quality metrics will yield the best results, but when real data is mixed in, that correlation no longer exists. To improve the results, the generative model type and its architecture are the critical factors to focus on. Diffusion models work best with transfer learning, while GAN-generated samples should be used for augmentation. In addition, transformer-based generative models need no image filtering, whereas those derived from CNNs must be selected through the realism score. 

Therefore, we demonstrate that synthetic images can improve a downstream task's performance. To fully leverage synthetic data, future work should prioritize ensuring its reliability in more limited datasets and exploring its application to diverse bioimage modalities and tasks.


\section*{Acknowledgements}
This research has been supported by the Elkartek Programme, Basque Government (Spain) (BEREZ-IA, KK-2023/00012 and IKUN, KK-2024/00003).

%
%
\bibliographystyle{splncs04}
\bibliography{main}

\begin{thebibliography}{10}
\providecommand{\url}[1]{\texttt{#1}}
\providecommand{\urlprefix}{URL }
\providecommand{\doi}[1]{https://doi.org/#1}

\bibitem{Ahmed2022-ja}
Ahmed, A.A., Abouzid, M., Kaczmarek, E.: Deep learning approaches in histopathology. Cancers (Basel)  \textbf{14}(21) (Oct 2022)

\bibitem{betzalel2022study}
Betzalel, E., Penso, C., Navon, A., Fetaya, E.: A study on the evaluation of generative models (2022)

\bibitem{budach2022effects}
Budach, L., Feuerpfeil, M., Ihde, N., Nathansen, A., Noack, N., Patzlaff, H., Naumann, F., Harmouch, H.: The effects of data quality on machine learning performance (2022)

\bibitem{dhariwal2021diffusion}
Dhariwal, P., Nichol, A.Q.: Diffusion models beat {GAN}s on image synthesis. In: NeurIPS (2021)

\bibitem{10.1001/jama.2017.14585}
Ehteshami~Bejnordi, B., Veta, M., Johannes~van Diest, P., van Ginneken, B., Karssemeijer, N., Litjens, G., van~der Laak, J.A.W.M., , the CAMELYON16~Consortium: {Diagnostic Assessment of Deep Learning Algorithms for Detection of Lymph Node Metastases in Women With Breast Cancer}. JAMA  \textbf{318}(22),  2199--2210 (2017). \doi{10.1001/jama.2017.14585}

\bibitem{Frid_Adar_2018}
Frid-Adar, M., Diamant, I., Klang, E., Amitai, M., Goldberger, J., Greenspan, H.: Gan-based synthetic medical image augmentation for increased cnn performance in liver lesion classification. Neurocomputing  \textbf{321},  321–331 (2018). \doi{10.1016/j.neucom.2018.09.013}

\bibitem{Gonzales2023-mr}
Gonzales, A., Guruswamy, G., Smith, S.R.: Synthetic data in health care: A narrative review. PLOS Digit Health  \textbf{2}(1),  e0000082 (Jan 2023)

\bibitem{goodfellow2014generative}
Goodfellow, I., Pouget-Abadie, J., Mirza, M., Xu, B., Warde-Farley, D., Ozair, S., Courville, A., Bengio, Y.: Generative adversarial nets. In: NeurIPS (2014)

\bibitem{Gurcan2009-ig}
Gurcan, M.N., Boucheron, L.E., Can, A., Madabhushi, A., Rajpoot, N.M., Yener, B.: Histopathological image analysis: a review. IEEE Rev Biomed Eng  \textbf{2},  147--171 (Oct 2009)

\bibitem{He_2016_CVPR}
He, K., Zhang, X., Ren, S., Sun, J.: Deep residual learning for image recognition. In: CVPR (2016)

\bibitem{heusel2018gans}
Heusel, M., Ramsauer, H., Unterthiner, T., Nessler, B., Hochreiter, S.: Gans trained by a two time-scale update rule converge to a local nash equilibrium. In: NeurIPS (2017)

\bibitem{DBLP:journals/corr/abs-2006-11239}
Ho, J., Jain, A., Abbeel, P.: Denoising diffusion probabilistic models. In: NeurIPS (2020)

\bibitem{Hossain2023}
Hossain, M.S., Armstrong, L.J., Abu-Khalaf, J., Cook, D.M.: The segmentation of nuclei from histopathology images with synthetic data. Signal, Image and Video Processing  \textbf{17}(7),  3703--3711 (2023). \doi{10.1007/s11760-023-02597-w}

\bibitem{Hou2015PatchBasedCN}
Hou, L., Samaras, D., Kurc, T.M., Gao, Y., Davis, J.E., Saltz, J.H.: Patch-based convolutional neural network for whole slide tissue image classification. In: CVPR (2016). \doi{10.1109/CVPR.2016.266}

\bibitem{kazeminia2019gans}
Kazeminia, S., Baur, C., Kuijper, A., {van Ginneken}, B., Navab, N., Albarqouni, S., Mukhopadhyay, A.: Gans for medical image analysis. Artificial Intelligence in Medicine  \textbf{109},  101938 (2020). \doi{https://doi.org/10.1016/j.artmed.2020.101938}

\bibitem{jimaging9040081}
Kebaili, A., Lapuyade-Lahorgue, J., Ruan, S.: Deep learning approaches for data augmentation in medical imaging: A review. Journal of Imaging  \textbf{9}(4) (2023). \doi{10.3390/jimaging9040081}

\bibitem{kingma2022autoencoding}
Kingma, D.P., Welling, M.: Auto-encoding variational bayes (2022), \url{https://arxiv.org/abs/1312.6114}

\bibitem{gan_mode}
Kossale, Y., Airaj, M., Darouichi, A.: Mode collapse in generative adversarial networks: An overview. In: ICOA (2022). \doi{10.1109/ICOA55659.2022.9934291}

\bibitem{HistoGAN}
Krause, J., Grabsch, H.I., Kloor, M., Jendrusch, M., Echle, A., Buelow, R.D., Boor, P., Luedde, T., Brinker, T.J., Trautwein, C., Pearson, A.T., Quirke, P., Jenniskens, J., Offermans, K., van~den Brandt, P.A., Kather, J.N.: Deep learning detects genetic alterations in cancer histology generated by adversarial networks. The Journal of Pathology  \textbf{254}(1),  70--79 (2021). \doi{https://doi.org/10.1002/path.5638}

\bibitem{ktena_generative_2024}
Ktena, I., Wiles, O., Albuquerque, I., Rebuffi, S.A., Tanno, R., Roy, A.G., Azizi, S., Belgrave, D., Kohli, P., Cemgil, T., Karthikesalingam, A., Gowal, S.: Generative models improve fairness of medical classifiers under distribution shifts. Nature Medicine  \textbf{30}(4) (2024). \doi{10.1038/s41591-024-02838-6}

\bibitem{basic_pathology}
Kumar, V., Abbas, A.K., Aster, J.C., Perkins, J.A.: Robbins basic pathology. Elsevier (2018)

\bibitem{kynkäänniemi2023role}
Kynk{\"a}{\"a}nniemi, T., Karras, T., Aittala, M., Aila, T., Lehtinen, J.: The role of imagenet classes in fr\'echet inception distance. In: ICLR (2023)

\bibitem{kynkäänniemi2019improved}
Kynk\"{a}\"{a}nniemi, T., Karras, T., Laine, S., Lehtinen, J., Aila, T.: Improved precision and recall metric for assessing generative models. In: NeurIPS (2019)

\bibitem{bioengineering10111289}
Labrada, A., Barkana, B.D.: A comprehensive review of computer-aided models for breast cancer diagnosis using histopathology images. Bioengineering  \textbf{10}(11) (2023). \doi{10.3390/bioengineering10111289}

\bibitem{lee2021selfdiagnosing}
Lee, J., Kim, H., Hong, Y., Chung, H.W.: Self-diagnosing gan: Diagnosing underrepresented samples in generative adversarial networks. In: NeurIPS (2021)

\bibitem{DBLP:journals/corr/LiuGNDKBVTNCHPS17}
Liu, Y., Gadepalli, K., Norouzi, M., Dahl, G.E., Kohlberger, T., Boyko, A., Venugopalan, S., Timofeev, A., Nelson, P.Q., Corrado, G.S., Hipp, J.D., Peng, L., Stumpe, M.C.: Detecting cancer metastases on gigapixel pathology images. CoRR  (2017), \url{http://arxiv.org/abs/1703.02442}

\bibitem{mirza2014conditional}
Mirza, M., Osindero, S.: Conditional generative adversarial nets. CoRR  (2014)

\bibitem{mo2019mining}
Mo, S., Kim, C., Kim, S., Cho, M., Shin, J.: Mining gold samples for conditional gans. In: NeurIPS (2019)

\bibitem{ddpm_medicine}
Moghadam, P., Dalen, S.V., Martin, K.C., Lennerz, J., Yip, S., Farahani, H., Bashashati, A.: A morphology focused diffusion probabilistic model for synthesis of histopathology images. In: WACV. IEEE Computer Society (2023). \doi{10.1109/WACV56688.2023.00204}

\bibitem{ddpm_beat_gans}
Müller-Franzes, G., Niehues, J.M., Khader, F., Arasteh, S.T., Haarburger, C., Kuhl, C., Wang, T., Han, T., Nolte, T., Nebelung, S., Kather, J.N., Truhn, D.: A multimodal comparison of latent denoising diffusion probabilistic models and generative adversarial networks for medical image synthesis. Scientific Reports  \textbf{13}(1) (2023). \doi{10.1038/s41598-023-39278-0}

\bibitem{peebles2023scalable}
Peebles, W., Xie, S.: Scalable diffusion models with transformers. In: ICCV (2023)

\bibitem{perez2017effectiveness}
Perez, L., Wang, J.: The effectiveness of data augmentation in image classification using deep learning (2017)

\bibitem{ramchandran2022learning}
Ramchandran, S., Tikhonov, G., Lönnroth, O., Tiikkainen, P., Lähdesmäki, H.: Learning conditional variational autoencoders with missing covariates. PR  \textbf{147},  110113 (2024). \doi{https://doi.org/10.1016/j.patcog.2023.110113}

\bibitem{rombach2022highresolution}
Rombach, R., Blattmann, A., Lorenz, D., Esser, P., Ommer, B.: High-resolution image synthesis with latent diffusion models. In: CVPR (2022)

\bibitem{ronneberger2015unet}
Ronneberger, O., Fischer, P., Brox, T.: U-net: Convolutional networks for biomedical image segmentation. In: Navab, N., Hornegger, J., Wells, W.M., Frangi, A.F. (eds.) MICCAI (2015)

\bibitem{imagenet}
Russakovsky, O., Deng, J., Su, H., Krause, J., Satheesh, S., Ma, S., Huang, Z., Karpathy, A., Khosla, A., Bernstein, M., Berg, A.C., Fei-Fei, L.: {ImageNet} large scale visual recognition challenge. International Journal of Computer Vision  \textbf{115}(3),  211--252 (2015)

\bibitem{sohldickstein2015deep}
Sohl-Dickstein, J., Weiss, E., Maheswaranathan, N., Ganguli, S.: Deep unsupervised learning using nonequilibrium thermodynamics. In: ICML (2015)

\bibitem{10.1001/jamanetworkopen.2019.14645}
Tomita, N., Abdollahi, B., Wei, J., Ren, B., Suriawinata, A., Hassanpour, S.: {Attention-Based Deep Neural Networks for Detection of Cancerous and Precancerous Esophagus Tissue on Histopathological Slides}. JAMA Network Open  \textbf{2}(11),  e1914645--e1914645 (2019). \doi{10.1001/jamanetworkopen.2019.14645}

\bibitem{Veeling2018-qh}
Veeling, B.S., Linmans, J., Winkens, J., Cohen, T., Welling, M.: Rotation equivariant cnns for digital pathology. In: Frangi, A.F., Schnabel, J.A., Davatzikos, C., Alberola-L{\'o}pez, C., Fichtinger, G. (eds.) Medical Image Computing and Computer Assisted Intervention -- MICCAI 2018. pp. 210--218. Springer International Publishing, Cham (2018). \doi{10.1007/978-3-030-00934-2_24}

\bibitem{wang2016deep}
Wang, D., Khosla, A., Gargeya, R., Irshad, H., Beck, A.H.: Deep learning for identifying metastatic breast cancer (2016)

\bibitem{XUE2021101816}
Xue, Y., Ye, J., Zhou, Q., Long, L.R., Antani, S., Xue, Z., Cornwell, C., Zaino, R., Cheng, K.C., Huang, X.: Selective synthetic augmentation with histogan for improved histopathology image classification. Medical Image Analysis  \textbf{67},  101816 (2021). \doi{https://doi.org/10.1016/j.media.2020.101816}

\bibitem{gan_multi_attribute}
Ye, J., Xue, Y., Liu, P., Zaino, R., Cheng, K.C., Huang, X.: A multi-attribute controllable generative model for histopathology image synthesis. In: de~Bruijne, M., Cattin, P.C., Cotin, S., Padoy, N., Speidel, S., Zheng, Y., Essert, C. (eds.) MICCAI (2021)

\bibitem{yu2023diffusionbased}
Yu, X., Li, G., Lou, W., Liu, S., Wan, X., Chen, Y., Li, H.: Diffusion-based data augmentation for nuclei image segmentation. In: Greenspan, H., Madabhushi, A., Mousavi, P., Salcudean, S., Duncan, J., Syeda-Mahmood, T., Taylor, R. (eds.) MICCAI. Springer Nature Switzerland (2023)

\bibitem{cutmix}
Yun, S., Han, D., Chun, S., Oh, S.J., Yoo, Y., Choe, J.: Cutmix: Regularization strategy to train strong classifiers with localizable features. In: ICCV (2019). \doi{10.1109/ICCV.2019.00612}

\bibitem{mixupzhang2018}
Zhang, H., Cisse, M., Dauphin, Y.N., Lopez-Paz, D.: mixup: Beyond empirical risk minimization. In: ICLR (2018)

\end{thebibliography}
\end{document}